\documentclass[letterpaper,10 pt,conference]{ieeeconf}
\IEEEoverridecommandlockouts

\usepackage{times}

\let\labelindent\relax
\usepackage{enumitem}
\usepackage{url}

\usepackage[usenames, dvipsnames]{xcolor}
\usepackage{graphicx}
\usepackage{siunitx}
\usepackage{booktabs}
\usepackage{tabularx}
\usepackage{caption}
\usepackage{subcaption}
\usepackage{environ}
\usepackage{textcomp}
\usepackage{amsmath}
\usepackage{amssymb}
\usepackage{afterpage}
\usepackage{pifont}
\usepackage{soul}
\usepackage[backend=bibtex,
            hyperref=false,
            url=false,
            isbn=false,
            doi=false,
            backref=false,
            style=ieee,
            citestyle=numeric-comp,
            sorting=nyt,%
            block=none]{biblatex}
\renewcommand{\bibfont}{\small}
\newcommand{\algfullname}{Language-Model-based Commonsense Reasoning}
\newcommand{\algname}{LMCR}

\def \emptycom#1{\unskip}
\def \emptycomm#1{}

\newcolumntype{Y}{>{\centering\arraybackslash}X}
\newcolumntype{P}{>{\raggedleft\arraybackslash}p{.2in}}

\makeatletter
\environfinalcode{}
\NewEnviron{xsubcaption}[1][-1sp]
 {\let\xsc@theoptcaption\@empty
  \let\xsc@thecaption\@empty
  \let\xsc@thelabel\@empty
  \expandafter\xsc@checkcaption\BODY\caption\xsc@checkcaption
  \expandafter\xsc@checklabel\BODY\label\xsc@checklabel
  \begingroup\edef\x{\endgroup
    \noexpand\subcaptionbox
      \ifx\xsc@theoptcaption\@empty\else
        [\unexpanded\expandafter{\xsc@theoptcaption}]%
      \fi
      {\unexpanded\expandafter{\xsc@thecaption}%
       \ifx\xsc@thelabel\@empty\else
         \noexpand\label{\unexpanded\expandafter{\xsc@thelabel}}%
       \fi}
      \ifdim#1=-1sp %
      \else
        [#1]%
      \fi
      {\unexpanded{\renewcommand\caption[2][]{}\renewcommand{\label}[1]{}}%
       \unexpanded\expandafter{\BODY}}}\x
}

\long\def\xsc@checkcaption#1\caption#2\xsc@checkcaption{%
  \if\relax\detokenize{#2}\relax
    \expandafter\@gobble
  \else
    \expandafter\@firstofone
  \fi
  {\expandafter\xsc@getcaption\BODY\xsc@getcaption}%
}
\long\def\xsc@getcaption#1\caption#2#3\xsc@getcaption{%
  \if[\detokenize{#2}%
    \expandafter\@firstoftwo
  \else
    \expandafter\@secondoftwo
  \fi
  {\expandafter\xsc@getcaptionopt\BODY\xsc@getcaptionopt}%
  {\def\xsc@thecaption{#2}}%
}
\long\def\xsc@getcaptionopt#1\caption[#2]#3#4\xsc@getcaptionopt{%
  \def\xsc@theoptcaption{#2}%
  \def\xsc@thecaption{#3}%
}
\long\def\xsc@checklabel#1\label#2\xsc@checklabel{%
  \if\relax\detokenize{#2}\relax
    \expandafter\@gobble
  \else
    \expandafter\@firstofone
  \fi
  {\expandafter\xsc@getlabel\BODY\xsc@getlabel}%
}
\long\def\xsc@getlabel#1\label#2#3\xsc@getlabel{%
  \def\xsc@thelabel{#2}%
}
\makeatother

\title{Enabling Robots to Understand Incomplete Natural Language Instructions Using Commonsense Reasoning}

\author{Haonan Chen$^{1\ast}$, Hao Tan$^{1}$, Alan Kuntz$^{2}$, Mohit Bansal$^{1}$, Ron Alterovitz$^{1}$
\thanks{This research was supported in part by the U.S.\ National Science Foundation (NSF) under Award CCF-1533844, DARPA MCS Grant N66001-19-2-4031, Google Focused Research Award, and ARO-YIP Award W911NF-18-1-0336.}%
\thanks{$^{1}$Department of Computer Science,
        University of North Carolina at Chapel Hill, Chapel Hill, NC 27599, USA.
        \tt{\{haonanchen, haotan, mbansal, ron\}@cs.unc.edu}}%
\thanks{$^{2}$Robotics Center and School of Computing, University of Utah, Salt Lake City, UT 84112, USA.
\tt{adk@cs.utah.edu}}%
\thanks{$^{\ast}$To whom correspondence should be addressed.}
}

\begin{document}
\maketitle

\begin{abstract}
Enabling robots to understand instructions provided via spoken natural language would facilitate interaction between robots and people in a variety of settings in homes and workplaces. However, natural language instructions are often missing information that would be obvious to a human based on environmental context and common sense, and hence does not need to be explicitly stated. In this paper, we introduce Language-Model-based Commonsense Reasoning (LMCR), a new method which enables a robot to listen to a natural language instruction from a human, observe the environment around it, and automatically fill in information missing from the instruction using environmental context and a new commonsense reasoning approach. Our approach first converts an instruction provided as unconstrained natural language into a form that a robot can understand by parsing it into verb frames. Our approach then fills in missing information in the instruction by observing objects in its vicinity and leveraging commonsense reasoning. To learn commonsense reasoning automatically, our approach distills knowledge from large unstructured textual corpora by training a language model. Our results show the feasibility of a robot learning commonsense knowledge automatically from web-based textual corpora, and the power of learned commonsense reasoning models in enabling a robot to autonomously perform tasks based on incomplete natural language instructions.
\end{abstract}

\section{Introduction} %
\begin{figure}
\centering
\includegraphics[width=\linewidth]{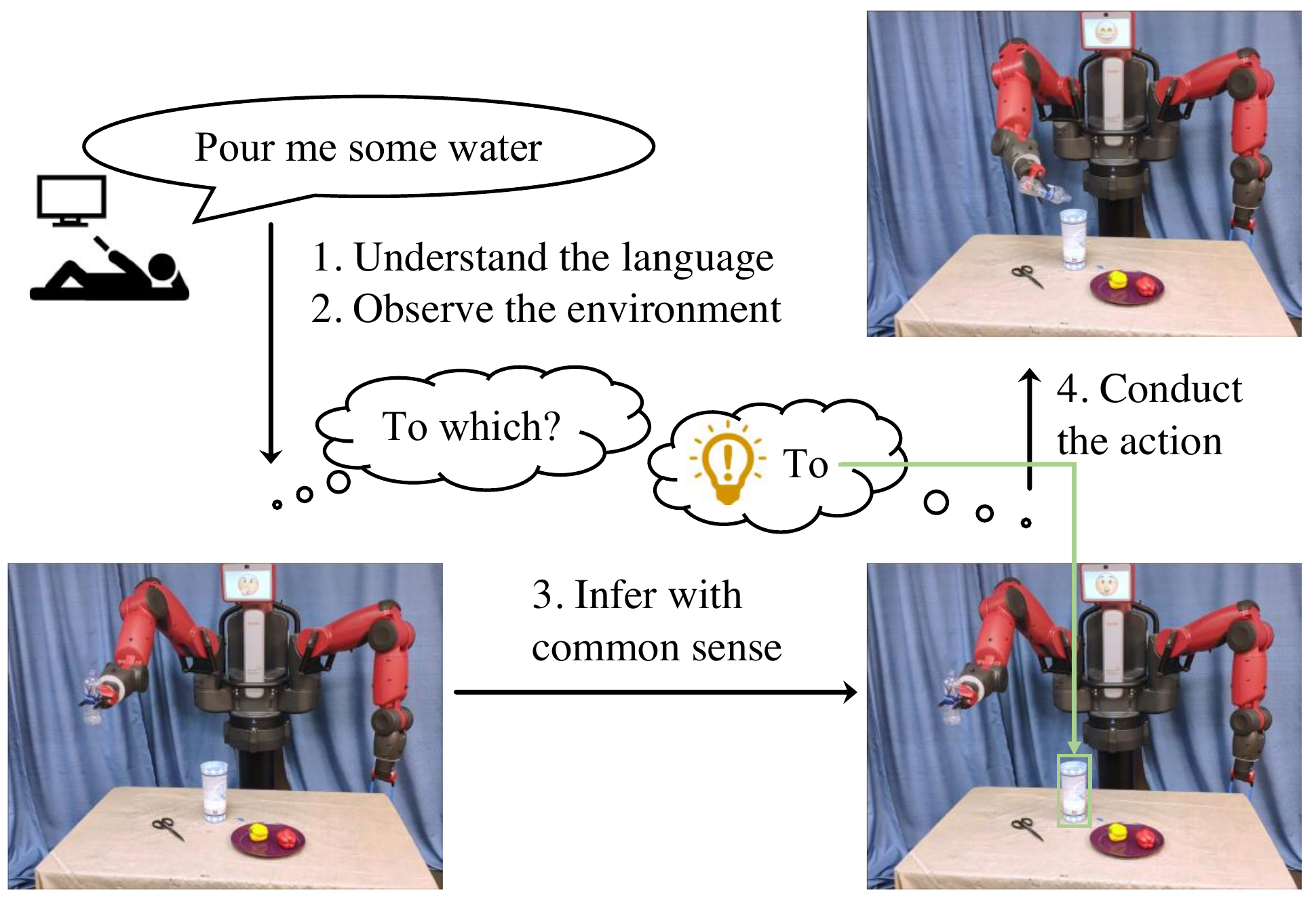}
\caption{\small\textbf{Common sense in instruction understanding.} A person gives an instruction ``pour me some water'' but the robot cannot carry out the action without knowing where to pour the water. After scanning the environment, the robot uses commonsense knowledge to determine the missing parameters and successfully perform the action.
}
\vspace{-0.6cm}
\label{fig:intro}
\end{figure}

\label{sec:introduction}
Natural language is inherently unstructured and often reliant on common sense to understand, which makes it challenging for robots to correctly and precisely interpret natural language.
Consider a scenario in a home setting in which a robot is holding a bottle of water and there are scissors, a plate, some bell peppers, and a cup on a table (see Fig.~\ref{fig:intro}).
A human gives an instruction, \emph{``pour me some water''}, to the robot.
This instruction is \emph{incomplete} from the robot's perspective since it does not specify where the water should be poured, but for a human, it might be obvious that the water should be poured into the cup.
A robot that has the common sense to automatically resolve such incompleteness in natural language instructions, just as humans do intuitively, will allow humans to interact with it more naturally and increase its overall usefulness.
To this end, we introduce \algfullname{} (\algname{}),
a new approach which enables a robot to listen to a natural language instruction from a human, observe the environment around it, automatically resolve missing information in the instruction, and then autonomously perform the specified task.

The core problem we are addressing is enabling a robot to understand incomplete natural language instructions with the help of commonsense reasoning, particularly handling cases in which an argument of the instruction's verb is missing. 
Solving this problem requires two steps: (1) identify if and how an instruction is incomplete, and (2) complete the instruction using knowledge of the objects in the robot's environment.

\begin{figure*}
\centering
\includegraphics[width=0.85\linewidth]{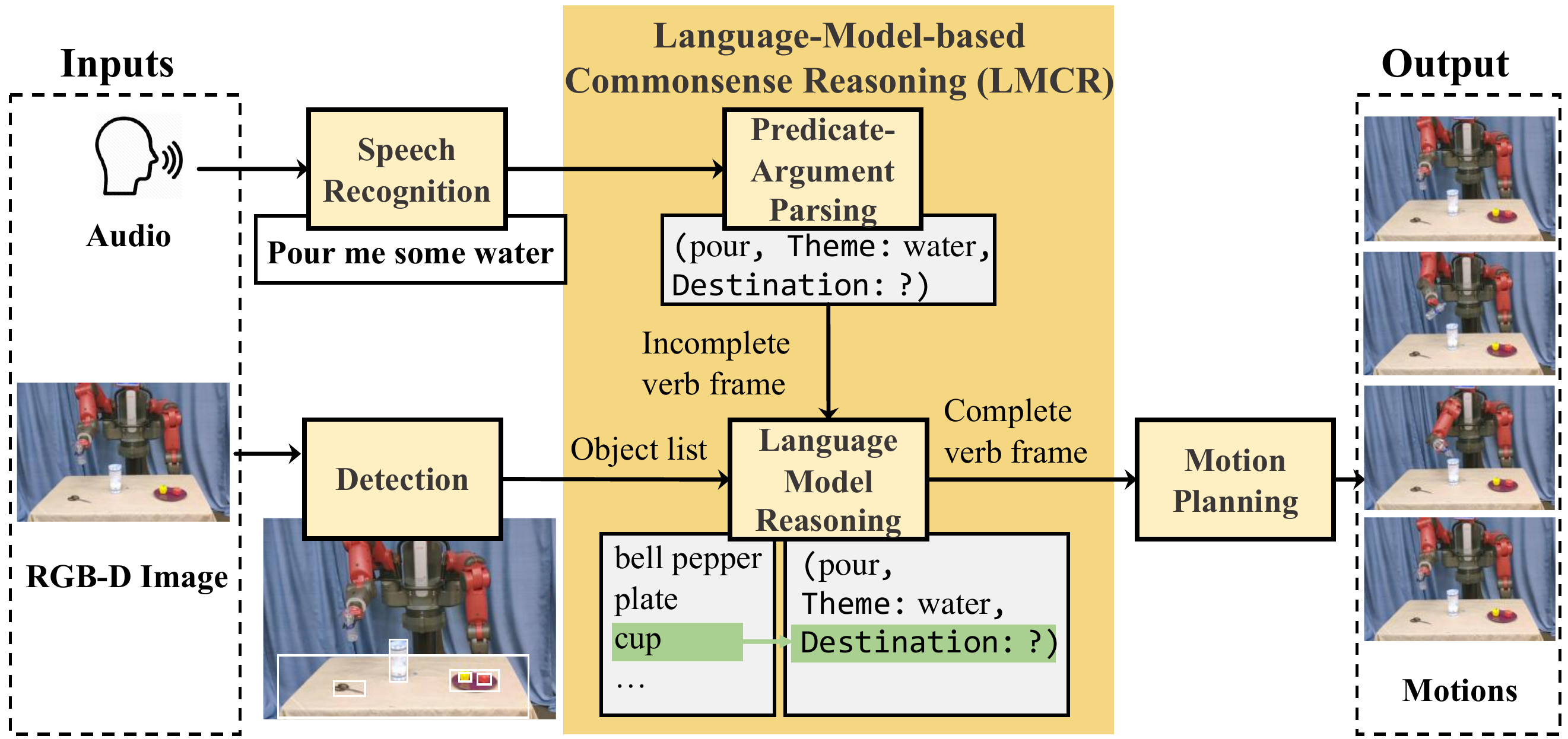}
\caption{\small\textbf{The components of a robot with \algname{}.}}
\vspace{-0.6cm}
\label{fig:system}
\end{figure*}

For the first step (identifying incomplete instructions), we parse the natural language instruction into a structured representation, referred to as a \emph{verb frame}. A verb frame is a tuple containing a predicate (i.e., a verb or verb phrase) and a set of semantic roles and their associated content \cite{jurafsky2014speech}. For example, \algname{} automatically parses the instruction \emph{``pour me some water''} to the verb frame (pour, {\ttfamily Theme}: water, {\ttfamily Destination}: {\ttfamily ?}), where ``pour'' is the predicate, ``water'' and ``{\ttfamily ?}'' are arguments that help complete the meaning of a predicate, and ``{\ttfamily Theme}'' and ``{\ttfamily Destination}'' are semantic roles which specify the underlying relationship between arguments and the predicate. The empty tag {\ttfamily ?} indicates that the argument of {\ttfamily Destination} is missing.
Under such a representation, incomplete instructions can be easily identified as not all roles in the verb frame are filled with content from the instruction.
The task of resolving incomplete instructions then becomes filling the missing role with objects in the environment.

For the second step (completing an incomplete instruction), we note that people are more likely to omit information from an instruction if it is obvious to the listener, so the correct role filler should be the one that yields a complete verb frame with the highest probability among all possible combinations. 
Inspired by this, LMCR uses a neural network based \emph{language model}, which acts as a probability distribution over sequences of words. After training on textual corpora containing descriptions of common household tasks, our language model is able to assign higher probabilities to candidate verb frames that correspond to more common complete instructions, such as (pour, {\ttfamily Theme}: water, {\ttfamily Destination}: {\ttfamily cup}). This language model, combined with limiting the missing arguments to objects in the robot's vicinity, enables the robot to automatically fill in missing information in natural language instructions via common sense.

We incorporate the above language understanding pipeline into a robot as shown in Fig.~\ref{fig:system}.
The robot gets instructions and environmental information via the Speech Recognition and Detection modules respectively, processes the inputs via \algname{}, and executes the specified task via the Motion Planning module.
A video of an \algname{}-enabled robot is provided in Supplementary Materials.
We also quantitatively evaluate \algname{} on a human-annotated dataset collected as part of this work.
We compile a novel dataset as existing datasets on commonsense reasoning such as \cite{chao2015mining,warren2015comprehending,pylkkanen2007meg} consist mostly of general-purpose verbs and nouns and are not aimed specifically at robot manipulation applications, making them unsuitable for evaluating our method.
In this work we focus on kitchen assistance tasks (e.g., blend, pour, sprinkle), but the same pipeline can be extended to other scenarios given relevant training data. 
The results show that incorporating commonsense knowledge via a language model approach enables a robot to understand and perform a task based on incomplete instructions, enabling more natural human-robot interaction.

\section{Related Work} %
\label{sec:discussion}
Reasoning using commonsense knowledge to understand incomplete natural language instruction has been studied in a variety of contexts. For example, Bolt et al.\ \cite{bolt1980put} presented a robotic system that could leverage deictic reference or pointing gestures to understand human instructions in a situated human-robot interaction setting.
Recent years have seen systems like Prac \cite{nyga2018cloud} and RoboBrain \cite{saxena2014robobrain} that have the ability to leverage world knowledge to understand natural language instructions. However, these systems tend to rely on graph-based knowledge representations. For example, Prac considered a similar commonsense reasoning problem as ours, aiming at inferring the most probable executable action in a given context, but the knowledge is encoded in a Prac knowledge base, which is constructed from manually annotated clauses found in natural language recipes.
\algname{}, by contrast, uses a neural network language model and is based on the intuition that world knowledge is implicitly encoded in textual corpora.
The idea is adapted from recent works in neural language models such as ELMo \cite{peters2018deep}, OpenAI GPT \cite{radford2018improving}, and BERT \cite{devlin2018bert}, using a pre-trained language model to improve the performance of various downstream applications, including commonsense reasoning.
These applications show that neural network language models are well suited to encoding and extracting knowledge that exists in large language corpora.

To understand natural language instructions, a robot has to extract a semantically meaningful representation of natural language and ground it to the perceptual elements and actions in its environment.
This process is referred to as language grounding \cite{matuszek2018IJCAI}.
Several approaches have been proposed for language grounding, which can be broadly divided into probabilistic models \cite{howard2014natural,hemachandra2015learning,paul2017grounding,paul2018efficient} and deterministic models \cite{thomas2012roboframenet,misra2016tell,misra2015environment,thomason2019improving}.
These approaches seek to find an intermediate representation in order to bridge natural language and machine commands.
To bridge this gap, the probabilistic models employ a probabilistic graphical model approach, while the deterministic models employ a frame-like structure.
Our proposed model falls into the deterministic model category.
However, the related work mentioned above does not consider grounding unstated concepts with the help of commonsense world knowledge.
Recently, due to the advancement of deep neural networks, several works use sequence learning and reinforcement learning to directly map text to actions, skipping the need for an intermediate representation of instructions \cite{janner2018representation,blukis2018mapping,shah2018follownet,wang2018reinforced,das2017embodied}. However, they either consider only navigation tasks, or a simple simulated environment, where the possible actions are limited.
In contrast, our method generalizes to any task domain as we can easily extend the set of our verb frame representations by adding more frames to our training corpus.

Affordance can be defined as knowledge of an object's functionality, and understanding affordances is crucial for a robot to recognize human activities, interact with the environment, and achieve its goals \cite{chao2015mining}.
Previous research on affordance can be primarily divided into two categories, namely, visual affordance and semantic affordance.
Our work is closely related to semantic affordance \cite{zhu2014reasoning,chao2015mining}, which seeks to model the possible actions that can be conducted on an object.
However, these works only model single verb-object pairs. We extend the dependency by using verb frames, which allows us to make inferences on object affordances conditioned on both the predicate and other roles.

\section{Method} %
\label{sec:method}
\vspace{-0.2cm}
\subsection{Problem Definition}

The robot receives a spoken instruction from the user as input. Our Speech Recognition module, shown in Fig.\ \ref{fig:system}, transcribes the audio of spoken language into text, which we specify as a sequence of $K$ tokens representing words, $W=\{w_1, w_2, \ldots, w_K\}$. We use Google Cloud API \cite{speech2text} for the transcription. The robot also receives input from its RGB-D sensors. Our Detection module in Fig.\ \ref{fig:system} detects instances of certain classes of objects and their positions in the input RGB-D image. 
This module can be implemented by an object detector, such as Mask R-CNN \cite{he2017mask}.
The Detection module outputs a list of relevant objects $\mathcal{O}$ in the vicinity of the robot, along with their associated positions.

We represent actions that the robot can perform using verb frames. 
Following the convention in the frame semantic parsing literature \cite{das2014frame,hermann2014semantic}, we define a verb frame as $f=(v,r_1,a_1,\dots, r_N,a_N)$, where $v$ denotes the predicate and $r_i$ and $a_i$ denote the $i$'th role and its argument, respectively. 
The predicate $v \in \mathcal{V}$ represents an action, where $\mathcal{V}$ is the set of actions that the robot can perform (e.g., ``pour'', ``brush'', as summarized in the left column of Table \ref{tab:res} for our robot). 
We focus on predicates (actions) that take $2$ arguments, so we simplify the verb frames to its two-argument specification $f=(v,r_1,a_1,r_2,a_2)$.
In our work, $r_i$ are drawn from a fixed, pre-defined set of role labels 
and are a function of the predicates.
At the same time, each $a_i$ is drawn from a fixed vocabulary (i.e., a set of words) $\mathcal{A}$. 
In our experiments, the labels (e.g., `apple', `banana') of detected objects $\mathcal{O}$ in a testing scenario come from this vocabulary $\mathcal{A}$.

The problem we want to study is to translate the possibly incomplete input instruction $W$ into a complete verb frame $f$ with all arguments filled in, while the detected object list $\mathcal{O}$ help with filling in the missing arguments.
Thus the robot's motion planner can execute this verb frame later.
Below, we first describe our approach for identifying missing arguments using verb frames (Sec.\ \ref{sec:IdentifyIncomplete}). 
We focus on the case where the human-provided instruction is missing one of the two roles.
We then introduce our approach to completing an incomplete verb frame using common sense via a neural-network based language model (Sec.\ \ref{sec:CompleteInstruction}), which will enable the robot to plan a motion to accomplish the desired task (Sec.\ \ref{sec:MotionPlanning}).

\subsection{Identifying Incomplete Instructions}
\label{sec:IdentifyIncomplete}

To identify if and how an instruction is incomplete, we parse the natural language instruction into a sequence of verb frames. The Predicate-Argument Parsing module in Fig.\ \ref{fig:system} takes the sequence of tokens $W$ and outputs a sequence of verb frames as input. We use an off-the-shelf semantic role labeling (SRL) model~\cite{he2017deep} to parse the sentence into verb frames, which provide us with a predicate-argument structure. Since some arguments may be missing from the instruction, we augment the vocabulary $\mathcal{A}$ to include an empty token, which is used to indicate a missing argument. 
Using parsed verb frames, an incomplete instruction can be identified as one having an empty token for one of its roles.
The problem of resolving an incomplete instruction then becomes filling the missing role with an object from the environment.

\subsection{Completing an Incomplete Instruction Using Common Sense}
\label{sec:CompleteInstruction}

Given an incomplete verb frame $f$ with one missing role and a list of objects $\mathcal{O}$ in the robot's environment, we formalize the task of commonsense reasoning as finding the most proper roll filler and outputting a complete verb frame.
This problem can be further treated as ranking a list of complete verb frames, as we can easily iterate over the object list $\mathcal{O}$ to create all possible candidate verb frames that are feasible in the current environment.
Thus, we implement commonsense reasoning as a scoring function $g(f)$ 
 where $f$ is a complete verb frame. 
 And from the list of candidate verb frames we pick the one with the highest score as the output verb frame.
We refer to the score as a \emph{plausibility score}.
The job of the commonsense reasoning method is then to define the scoring function $g$.

To compute the plausibility score, we note that people are more likely to omit information from an instruction if it is obvious to the listener, so the correct role filler should be the one that yields a complete verb frame with the highest probability among all possible combinations. To this end, we use a language model (LM) whose goal is to predict the probability of a word sequence (we assume a word sequence with higher probability to appear is more plausible). A language model factorizes the probability according to the chain rule. 
Using $\{u_1, u_2, \dots, u_T\}$ to denote an entire sentence with $T$ tokens, the chain rule can be written as,
\begin{equation}
p(u_1, u_2, \dots, u_T) = \prod^{T}_{t=1}p(u_t| u_1,u_2,\dots,u_{t-1}),
\end{equation}
where $p(u_t|u_1, u_2, \ldots, u_{t-1})$ is the conditional probability of the word $u_t$ given the previous words. In the Language Model Reasoning module of our work, we model this conditional probability using a recurrent neural network (RNN) \cite{mikolov2010recurrent}.

Following the recent progress in the study of language models, we also tried other advanced pre-trained language models such as ELMo \cite{peters2018deep} and BERT \cite{devlin2018bert}.
However, we empirically did not find a significant difference between these different language models of our approach, so we take the simplest RNN-based LM as our model here.

Note that the language model operates on a sequence of words, but verb frames are a structured representation of language. We thus need to serialize the candidate complete verb frames into a sequence of words, a process known as \emph{linearization} \cite{filippova2009tree,konstas2017neural}.
We propose two \emph{linearization} methods in this work.
The first is to concatenate the predicate and all arguments directly, i.e., to treat $(v, a_1, a_2)$ as a sequence. This results in unnatural sounding word sequences. The second is to make a more natural sentence from the frame using a rule-based approach.
With these two approaches, (pour, {\ttfamily Theme:} water, {\ttfamily Destination:} cup) is converted to \emph{pour water cup} with the former approach and \emph{pour water to the cup} with the latter one. We refer to the LM trained and tested with the former approach as frame-based LM and the latter one as sentence-based LM.
For both, the sequence format needs to be consistent during training and inference to get the best performance.
As our training corpus contains natural language sentences, we can use them to train the sentence-based LM directly, while frame-based LM requires predicate-argument parsing on the entire training corpus as a pre-processing step.

\subsection{Motion Planning for a Complete Verb Frame}
\label{sec:MotionPlanning}

The motion planner takes as input a complete verb frame $f$ and the positions of relevant objects in $\mathcal{O}$ and computes a motion for the robot that executes the task specified by the verb frame.
For each $v \in \mathcal{V}$, we define a motion planner parameterized by its arguments.
In our implementation, each motion planner is defined by a series of waypoints for the end-effector. Each waypoint is defined in a coordinate system relative to the positions of a task-relevant object in $\mathcal{O}$ \cite{Bowen2015_TASE}, which enables the robot to plan motions that are robust to the movement of the objects in the environment.
Reaching these waypoints in sequence executes the action.
We use the motion planning toolkit MoveIt! \cite{moveit2018} to compute the movement of the robot arm given the relative waypoints.

\subsection{Comparison Methods for Commonsense Reasoning Evaluation}
As described above, \algname{} gives a score to each complete verb frame $f=(v, r_1, a_1, r_2, a_2)$ in a generated list, and the frame with the highest score is chosen as the output.
We use $g(f)$ to denote the scoring function.
In Sec.~\ref{sec:res_eff}, we compare the scoring function of our method \algname{} against those of co-occurrence, Word2Vec, and ConceptNet, described below.

\paragraph{Co-occur} 
This is the shorthand of ``co-occurrence''. Chao et al. \cite{chao2015mining} used co-occurrence in a textual corpus to determine the relatedness of a verb-object pair. We extend this to determine the relatedness of a verb frame, which is defined as
\begin{equation}
g_\mathrm{cooccur}(f) = \mathrm{cooccur}(v, a_2) + \mathrm{cooccur}(a_1, a_2),
\end{equation}
where $\mathrm{cooccur}(x, y)$ denotes the total normalized co-occurrence score of $x$ and $y$ in the training text corpora. This is computed by $\mathrm{count}(x, y) / (\mathrm{count}(x)*\mathrm{count}(y))$ where $\mathrm{count}(x)$ and $\mathrm{count}(y)$ are the occurrences of $x$ and $y$ individually in the corpus and $\mathrm{count}(x, y)$ is the count of $x$ and $y$ co-occurring in the same sentence.

\paragraph{Word2Vec}
Chao et al. \cite{chao2015mining} also used Word2Vec as one of their affordance mining methods. Similarly, we extend it to work on verb frames by defining the scoring function as
\begin{equation}
g_\mathrm{word2vec}(f)=-(\mathrm{dist}(v, a_2) + \mathrm{dist}(a_1, a_2)),
\end{equation}
where $\mathrm{dist}(x, y)$ denotes the Euclidean distance of word embeddings of $x$ and $y$.
We use GloVe embeddings \cite{pennington2014glove} for this comparison.

\paragraph{ConceptNet}
Systems like PRAC and RoboBrain use knowledge graphs and conduct probabilistic inference on the graph for instruction completion. Similarly, we use ConceptNet \cite{speer2017conceptnet}, which is a large scale common sense knowledge graph. We use the relatedness score provided by the ConceptNet API \cite{conceptnet5api}, and compute the score for a frame $f$ as
\begin{equation}
g_\mathrm{ConceptNet}(f)=\mathrm{rel}(v, a_2) + \mathrm{rel}(a_1, a_2),
\end{equation}
where $\mathrm{rel}(x, y)$ denotes the ConceptNet relatedness score \cite{speer2017conceptnet} of $x$ and $y$.

\section{Datasets} %
\label{sec:results}
\paragraph{Training Data for the Language Model}
The training data for \algname{}'s language model comes from textual corpora, which can be treated as the knowledge source of the method.
We use YouCook2 \cite{zhou2017procnets} and Now You're Cooking (NYC) \cite{nyc2013} as training corpora.
YouCook2 is a large instructional video dataset designed to facilitate video captioning research. The cooking steps for each video are annotated with temporal boundaries and described by imperative English sentences, resulting in around $14,000$ raw descriptions of cooking actions. NYC contains over $150,000$ recipes, each containing a step-by-step description of how to execute the recipe. Although NYC is much larger in size, it contains unrelated information such as ingredient lists and comments, which is more similar to what we can get directly by crawling web data.
In our experiment, we deliberately keep this extraneous information in order to determine if the language model can distill commonsense knowledge required in human-robot interaction from noisy textual corpora.

\begin{figure}
\centering
\scriptsize
\begin{minipage}[b]{\linewidth}
\includegraphics[width=\linewidth]{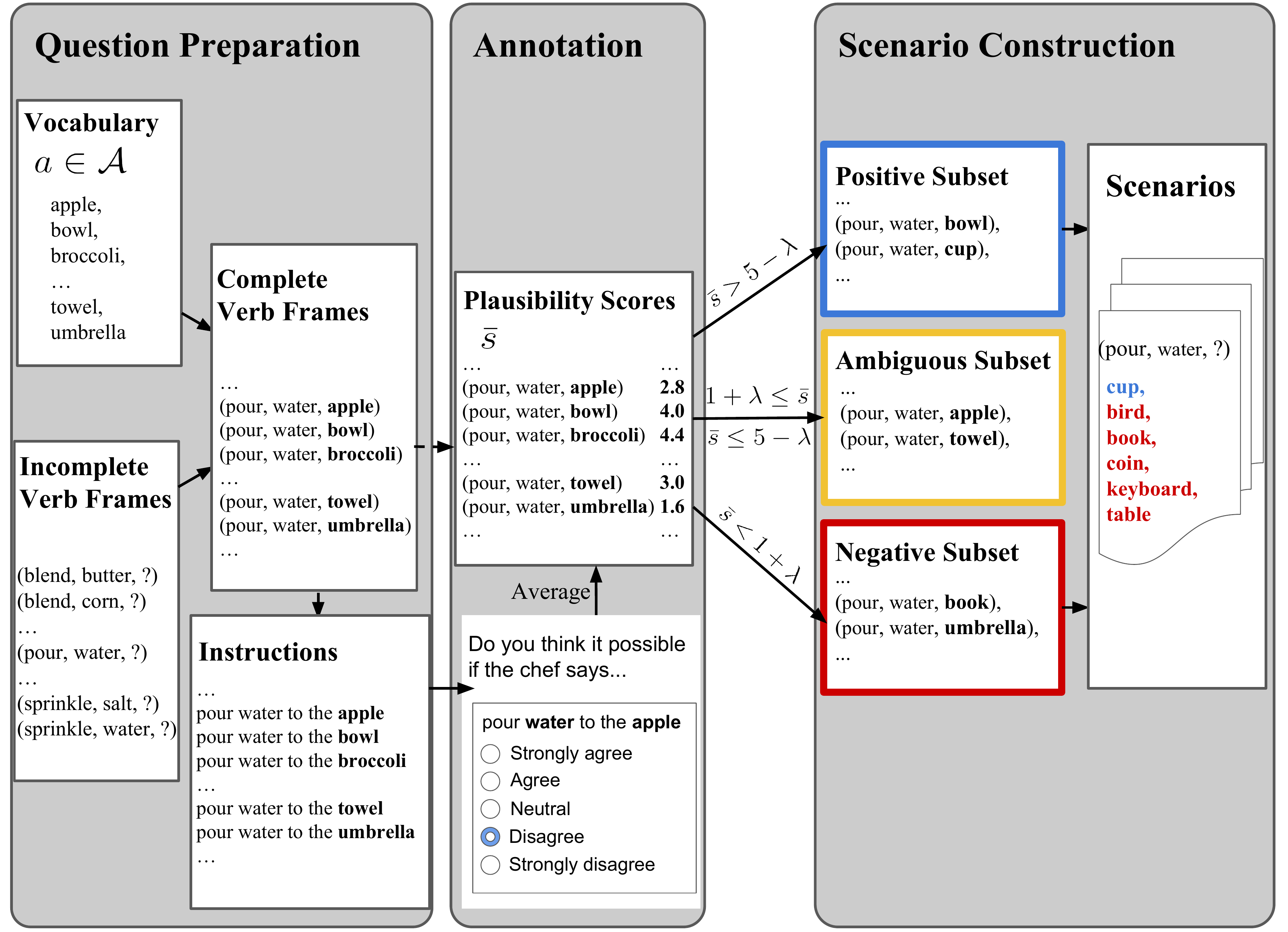}
\end{minipage}
\caption{\small \textbf{Human judgment collection and scenario preparation.
}
}
\label{fig:eff}
\end{figure}

\begin{table}
\centering
\small
\caption{Overall accuracy and accuracy per predicate for $\lambda=1.0$ and $k=6$.}
\label{tab:res}
\begin{tabularx}{0.99\linewidth}{l YYYYYp{0.1cm}}
\toprule
{} &  \textbf{Random} &  & \textbf{Word2Vec} & \multicolumn{3}{r}{\textbf{LMCR (Ours)}} \\
{} & & \textbf{Cooccur} & & \multicolumn{1}{X}{\textbf{ConceptNet}} & & \\
\midrule
blend    &  0.18 & 0.46 & 0.66 & 0.66 & \textbf{0.88} & \\
brush    &  0.18 & 0.58 & 0.50 & 0.26 & \textbf{0.86} & \\
dip      &   0.10 & 0.42 & 0.20 & 0.40 & \textbf{0.50} & \\
dump     &  0.24 & \textbf{0.68} & 0.64 & 0.58 & 0.54 & \\
fill     &   0.20 & 0.62 & 0.46 & 0.42 & \textbf{0.94} & \\
fry      &  0.18 & \textbf{0.86} & 0.48 & 0.62 & 0.66 & \\
heat     &  0.06 & \textbf{0.74} & 0.40 & 0.70 & 0.64 & \\
pour     &  0.08 & \textbf{0.74} & 0.56 & 0.52 & 0.60 & \\
rub      &  0.04 & \textbf{0.54} & 0.38 & 0.40 & 0.50 & \\
season   &  0.16 & 0.58 & 0.50 & \textbf{0.84} & \textbf{0.84} & \\
sprinkle &   0.20 & 0.58 & 0.54 & 0.36 & \textbf{0.78} & \\
\midrule
Overall    &  0.15  & 0.62 & 0.48 & 0.52 & \textbf{0.70} & \\
\bottomrule
\end{tabularx}
\vspace{-0.2cm}
\end{table}
\paragraph{Testing Data Based on Human Judgments}

In order to quantitatively evaluate \algname{}'s commonsense reasoning for robotic assistance instructions, we created a new human-generated dataset, since existing datasets on commonsense reasoning~\cite{chao2015mining,warren2015comprehending,pylkkanen2007meg} are not specific to our domain of filling in missing information in instructions for robotic assistance tasks. We show our data collection process in Fig.~\ref{fig:eff}. We provided human annotators on Amazon Mechanical Turk (AMT) \cite{amt} with sentences representing complete verb frames and asked them to give a plausibility rating for each of them, scaling from 1 (most implausible) to 5 (most plausible) (see the figure for examples). We collected 5 annotations for all complete verb frames in our dataset. 

In our experiments, for each predicate we split the verb frames into positive, ambiguous, and negative subsets using a \emph{plausibility threshold} $\lambda$.
Namely, for a complete verb frame with an average plausibility rating $\bar{s}$, it is included in the positive subset if $\bar{s} > 5-\lambda$, the negative subset if $\bar{s} < 1 + \lambda$, and the ambiguous subset otherwise. 
We then randomly pick one frame from the positive and $k-1$ frames from the negative subset (we restrict one correct answer in a test scenario for the convenience of evaluation), keeping the predicate and one of the arguments the same and varying the other argument. In this way, we can construct a test scenario with $k$ candidates, where one of them is plausible based on the human annotation.

In our experiments, we vary the plausibility threshold $\lambda$ and the number of candidates $k$ to create test scenarios with different difficulties. A larger $\lambda$ brings more ambiguous frames (which even humans are not sure about their plausibility) into the positive and negative subsets. A larger $k$ introduces more candidates in a single scenario. In both cases, the test scenarios become more challenging.

\section{Results}
\label{sec:res_eff}
\begin{figure*}
\centering
\begin{minipage}[b]{0.4\linewidth}
\includegraphics[width=\linewidth, trim=2cm 0.5cm 1cm 0]{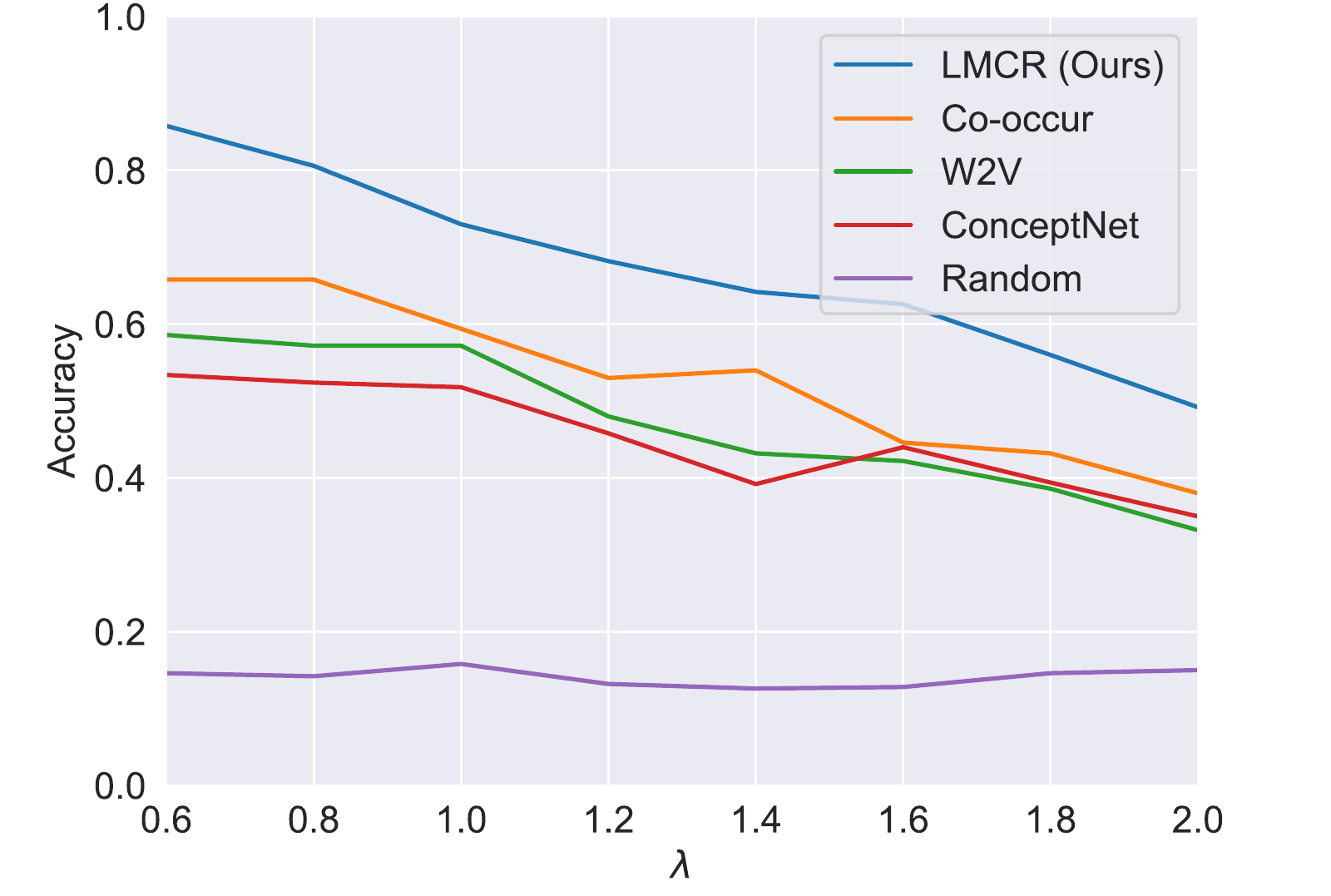}
\caption*{\scriptsize\textbf{A}}
\end{minipage}\hspace{1cm}
\begin{minipage}[b]{0.4\linewidth}
\includegraphics[width=\linewidth, trim=2cm 0.5cm 1cm 0]{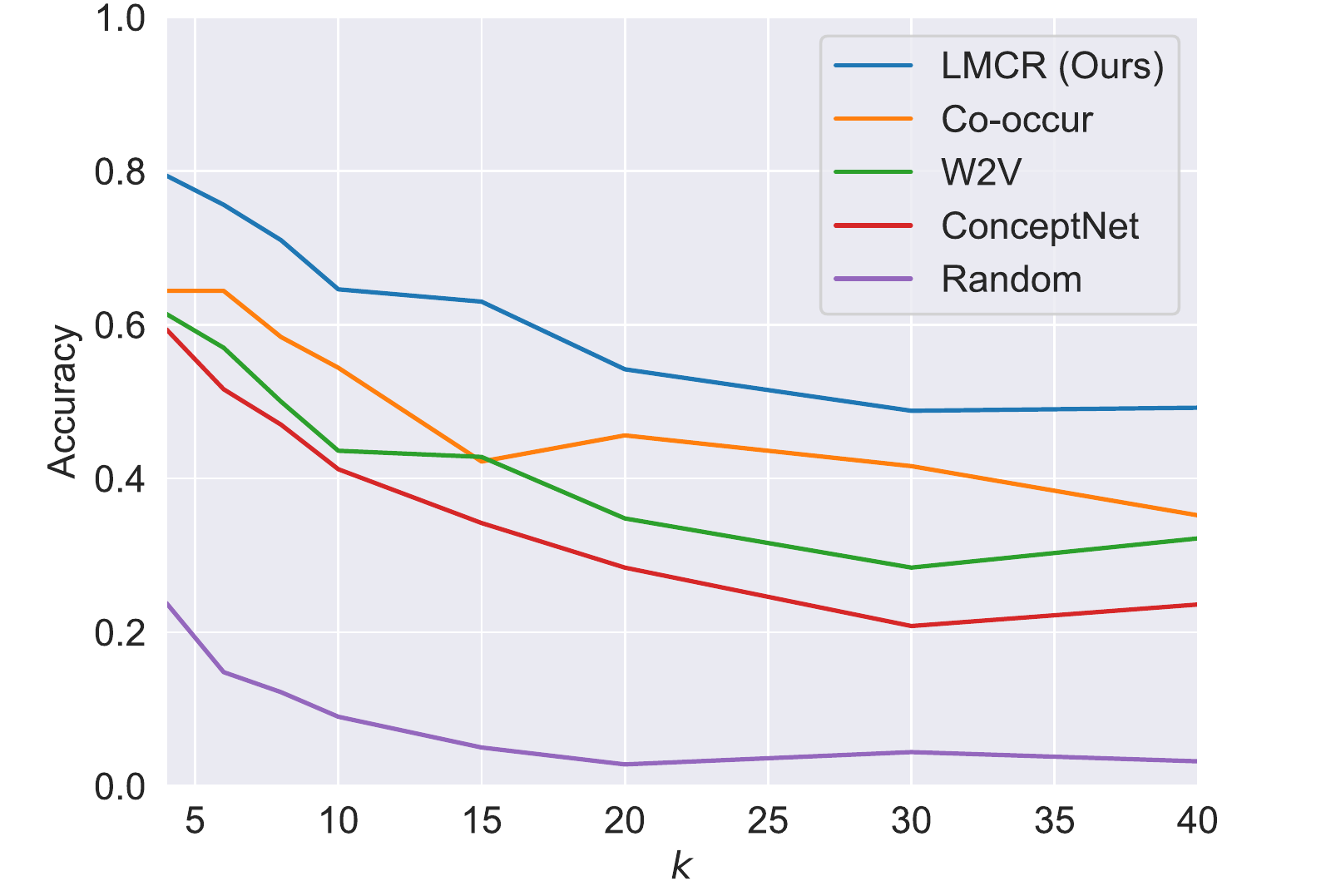}
\caption*{\scriptsize\textbf{B}}
\end{minipage}\\
\begin{minipage}[b]{0.4\linewidth}
\includegraphics[width=\linewidth, trim=2cm 0.5cm 1cm 0cm]{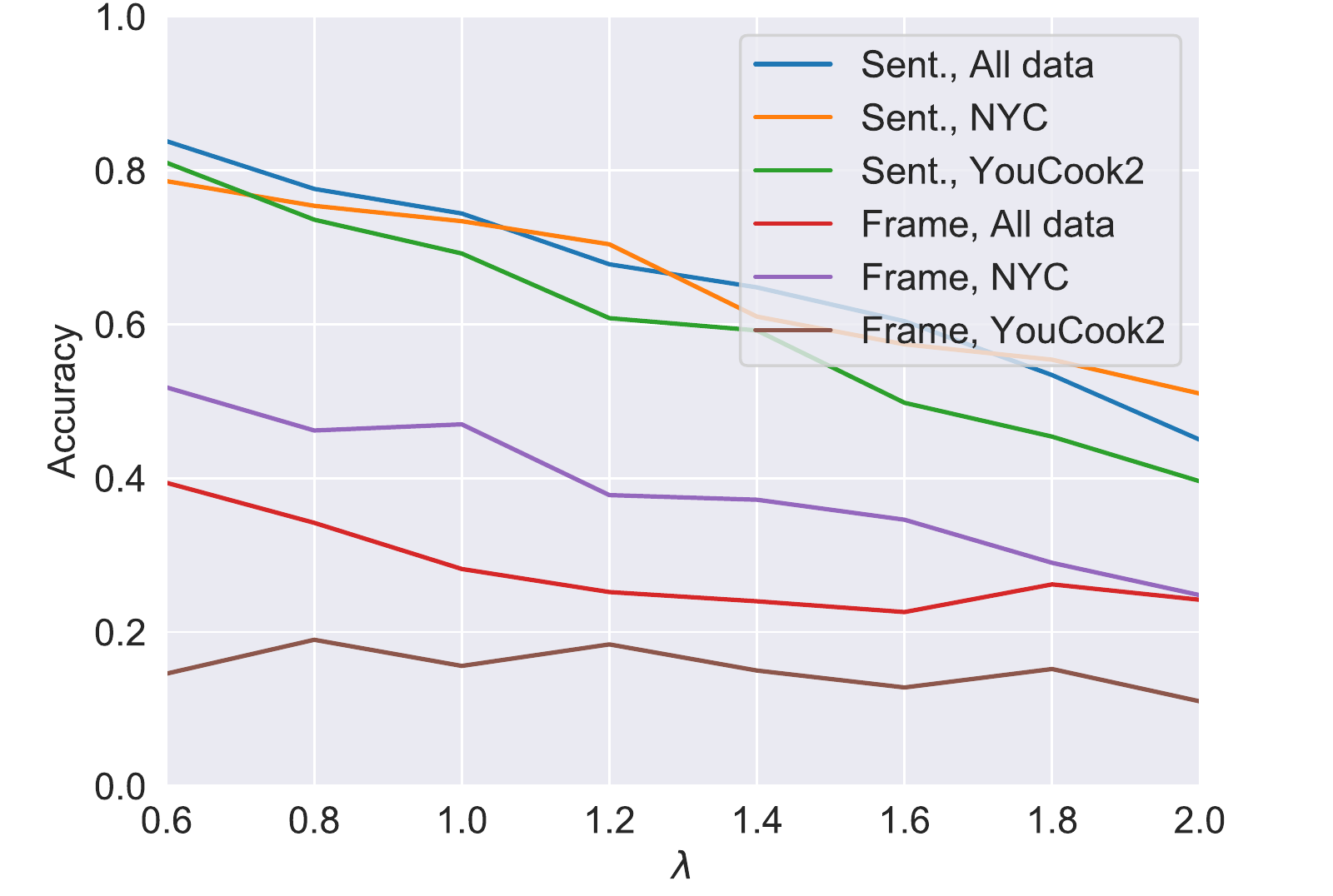}
\caption*{\scriptsize\textbf{C}}
\end{minipage}\hspace{1cm}
\begin{minipage}[b]{0.4\linewidth}
\includegraphics[width=\linewidth, trim=2cm 0.5cm 1cm 0cm]{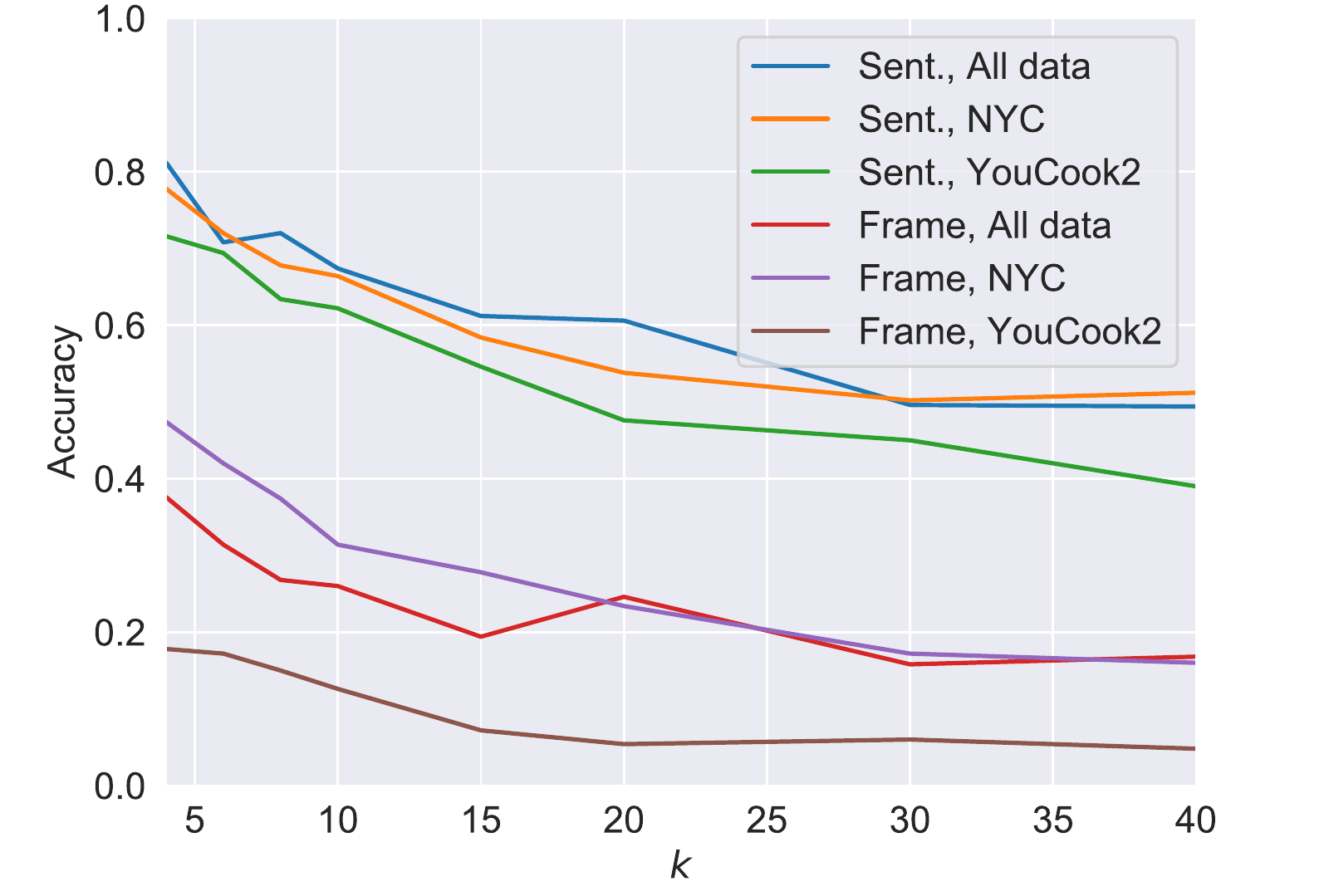}
\caption*{\scriptsize\textbf{D}}
\end{minipage}
\caption{\small\textbf{Accuracy results for \algname{}.}  
We evaluate the ability of \algname{} to correctly fill in missing information for scenarios of varying difficulty, by changing the plausibility threshold $\lambda$ and number of objects per scenario $k$. (\textbf{A}) Overall accuracy with $\lambda$ changing from $0.6$ to $2.0$ and $k=6$ for \algname{} and baseline methods. (\textbf{B}) Overall accuracy with $k$ changing from $4$ to $40$ and $\lambda=1.0$ for \algname{} and baseline methods. (\textbf{C}) Overall accuracy with $\lambda$ changing from $0.6$ to $2.0$ and $k=6$ for different language model training settings. (\textbf{D}) Overall accuracy with $k$ changing from $4$ to $40$ and $\lambda=1.0$ for different language model training settings. There are 500 test scenarios in all settings.
\vspace{-0.75cm}
}
\label{fig:eff_res}
\end{figure*}

\paragraph{Comparison with Other Methods} Based on the collected human judgment dataset, we compare the proposed \algname{} approach\footnote{The language model here is the sentence-based LM trained on both the YouCook2 and Now You're Cooking dataset.} with other baseline methods Co-occur, Word2Vec, and ConceptNet, described above, as well as with a uniform random choice (Random).
Each method defines a scoring function $g(f)$ given a complete verb frame $f$. Given a test scenario containing $k$ candidate verb frames, a successful prediction gives the highest score to the ground truth, namely the one sampled from the positive subset.
We consider 11 verbs listed in the leftmost column of Table~\ref{tab:res}, vary the plausibility threshold $\lambda$ and the number of objects in the list $k$ to create scenarios with various difficulties, and report the accuracy (success rate).
Table~\ref{tab:res} gives the overall and per predicate accuracy with $k=6$ and plausibility threshold $\lambda=1.0$, and Fig.~\ref{fig:eff_res}A and Fig.~\ref{fig:eff_res}B show the results when varying $\lambda$ and $k$ respectively.
\algname{} performs consistently better than other methods when considering all actions (predicates), for all variations of $k$ and $\lambda$, although some methods show better performance on specific individual predicates.
The results suggests that, overall, \algname{} better encodes the type of commonsense reasoning we are addressing in this work.

\paragraph{Comparison under Different Training Settings} We also compare several different ways of training the language model (LM) used by the Language Model Reasoning module of \algname{}.
To do so we train the LM with two different linearization strategies, namely, frame-based (Frame) and sentence-based (Sent.)\ linearization.
We also train the LM on different combinations of training corpora, YouCook2 data only (YouCook2), Now You're Cooking data only (NYC), and the combination of the two (All data).
Results for these comparisons are shown in Fig.~\ref{fig:eff_res}C and Fig.~\ref{fig:eff_res}D which vary $\lambda$ and $k$ respectively.
As shown in the results, sentence-based LM generally performs better than frame-based LM.
We suspect this is due to the fact that the former is end-to-end trained while the latter requires generating training data from an upstream predicate-argument parser, whose errors may propagate to the training process of the frame-based LM.
Also, the parser cannot generate a frame for predicates that are not in its verb vocabulary, even though these relatively rare predicates can be beneficial when learning others.
For example, ``scatter some salt on the beef'' would help with the learning for ``spread'' and ``sprinkle'' as they can be synonyms.
While the sentence-based LM can take advantage of this, the frame-based schema loses this information in the training data, since ``scatter'' is not among the 11 verbs we consider.
For sentence-based LM, the performance of the models trained with NYC and all
data (YouCook2+NYC) are similar, and both are better than the model trained on YouCook2 alone.
For frame-based LM, the all data yields the best performance.
Although NYC is noisier than YouCook2, the former still brings positive input to the language model training, since it is much larger than the latter.
This suggests that the language model is robust to the noise in the training data on the commonsense reasoning task considered in this paper.
These results demonstrate that a human annotated dataset, such as YouCook2, is not necessarily better than a recipe-based dataset, although it may still be helpful.
Based on the above analyses, we use the sentence-based LM trained on all data when comparing with the other commonsense reasoning approaches in the previous section.
\paragraph{Real Robot Experiment} We deploy \algname{} on a Baxter robot \cite{rethink2013baxter}, a research robotics platform with two 7 degree of freedom arms, and demonstrate its ability to successfully accomplish intended tasks given incomplete spoken instructions in different scenarios. A video of the LMCR-enabled robot in action is provided in Supplementary Materials.
\vspace{-.6em}
\section{Conclusion}
In this work, we presented a robot that can detect when a human instruction is incomplete and automatically resolve it by observing the environment and making inferences based on commonsense world knowledge. The use of a neural language model in capturing the commonsense knowledge allows us to leverage online textual corpora and train the model with little manual intervention. We demonstrate the effectiveness of our algorithm both by measuring the alignment with human judgments and on a physical robot. 
In future work, we plan to investigate the robustness of the entire system against error in each module, consider verb frames with a varying number of missing arguments, and use dialogue when LMCR cannot make a confident decision about filling in missing information.

\renewcommand*{\bibfont}{\footnotesize}
\begin{flushright}
\printbibliography %
\end{flushright}
\clearpage

\end{document}